\documentclass{article}
\usepackage{ijcai26}

\usepackage{times}
\usepackage{soul}
\usepackage{url}
\usepackage[hidelinks]{hyperref}
\usepackage[utf8]{inputenc}
\usepackage[small]{caption}
\usepackage{graphicx}
\usepackage{svg}
\usepackage{amsmath}
\usepackage{amssymb}
\usepackage{amsthm}
\usepackage{tabularx}
\usepackage{booktabs}
\usepackage{algorithm}
\usepackage{algorithmic}
\usepackage[switch]{lineno}
\usepackage{enumitem}
\usepackage{pgfplots}
\usepackage{cleveref}
\usepackage{wrapfig}
\usetikzlibrary{patterns}
\newcolumntype{R}{>{\raggedleft\arraybackslash}X}

\usepackage{xcolor}
\usepackage{chato-notes} 

\newcommand{\cn}{%
    \textcolor{red!60!black}{\textbf{\textsc{[Citation needed]}}}%
}
\newcommand{\citet}[1]{{\citeauthor{#1}~\shortcite{#1}}}

\newtheorem{example}{Example}

\newtheorem{definition}{Definition}
\newtheorem{remark}{Remark}

\title{The EpisTwin: A Knowledge Graph-Grounded Neuro-Symbolic Architecture for Personal AI}

\author{
Giovanni Servedio$^1$\and
Potito Aghilar$^1$\and
Alessio Mattiace$^1$\and\\
Gianni Carmosino$^1$\and
Francesco Musicco$^1$\and
Gabriele Conte$^1$\and\\
Vito Walter Anelli$^1$\and
Tommaso Di Noia$^1$\And
Francesco Maria Donini$^2$
\affiliations
$^1$Politecnico di Bari, 
$^2$Università degli Studi della Tuscia\\
\emails
\{g.conte12, a.mattiace, g.carmosino1\}@studenti.poliba.it,\\ \{potito.aghilar, giovanni.servedio, vitowalter.anelli, tommaso.dinoia\}@poliba.it\\
f.musicco@phd.poliba.it, donini@unitus.it
}

\begin{document}

\maketitle

\begin{abstract}
Personal Artificial Intelligence is currently hindered by the fragmentation of user data across isolated silos. While Retrieval-Augmented Generation offers a partial remedy, its reliance on unstructured vector similarity fails to capture the latent semantic topology and temporal dependencies essential for holistic sensemaking. We introduce EpisTwin, a neuro-symbolic framework that grounds generative reasoning in a verifiable, user-centric Personal Knowledge Graph.
EpisTwin leverages Multimodal Language Models to lift heterogeneous, cross-application data into semantic triples.
At inference, EpisTwin enables complex reasoning over the personal semantic graph via an agentic coordinator that combines Graph Retrieval-Augmented Generation with Online Deep Visual Refinement, dynamically re-grounding symbolic entities in their raw visual context.
We also introduce \textit{PersonalQA-71-100}, a synthetic benchmark designed to simulate a realistic user's digital footprint and evaluate EpisTwin’s performance. Our framework demonstrates robust results across a suite of state-of-the-art judge models, offering a promising direction for trustworthy Personal AI.
\end{abstract}


\section{Introduction}
The promise of Personal Artificial Intelligence is to function as a cognitive extension of the user~\cite{BalogK19}.
However, this vision is constrained by the fragmentation of user data~\cite{DBLP:journals/aiopen/SkjaevelandBBLL24,DBLP:journals/widm/ChakrabortyS23} across an archipelago of isolated applications (e.g., calendars, galleries, document stores). Lacking a unified semantic layer, these silos obscure the latent connections required for holistic sensemaking, preventing effective reasoning about the user's past, present, and future context.
Consider the cognitive burden involved in answering a natural, cross-domain request such as:

\begin{example}[Sarah call]
"Did Sarah Green call me before or after I arrived at work today?".
\normalfont Answering this seemingly simple question may require complex reasoning on personal data. For example, if the user was late for work, the system may need to check today's call log, the start time of the recurring “Work” calendar event, and even notes, photos, or messages that suggest his actual arrival time.
\end{example}

Current systems force the user to bridge these silos manually. While Retrieval-Augmented Generation (RAG)~\cite{DBLP:conf/nips/LewisPPPKGKLYR020} mitigates this problem by grounding generation in retrieved data, it struggles with ``global" sensemaking, as it retrieves fragments based on local semantic overlap rather than the topological and temporal dependencies essential for personal reasoning. Furthermore, the reliance on opaque vector stores poses challenges for data sovereignty: the deterministic ``unlearning" of specific facts, a requirement under tightening regulations~[\citeyear{AIAct}], is computationally non-trivial in purely dense representations~\cite{DBLP:journals/tist/NguyenHRNLYN25}.

In this paper, we introduce \textbf{EpisTwin} (\textbf{Epist}emic \textbf{Twin}), an agentic framework that \textit{inverts the standard paradigm}: rather than treating the LLM as a probabilistic knowledge store, we use it as a structural architect to populate and reason over a \textbf{Personal Knowledge Graph (PKG)}. This decoupling yields an explicit, verifiable, symbolic structure in which knowledge deletions are deterministic.
EpisTwin aligns with Type~3 (Neuro~$\vert$~Symbolic) architecture in Kautz’s taxonomy~\cite{Kautz22}, defined by a cooperative relationship where the neural component acts as a coroutine rather than a mere subroutine. Initially, Type 3 was conceived as the neural transformation of non-symbolic feedback signals to symbolic inputs for symbolic reasoning. EpisTwin adapts this paradigm for inference-time symbolic and neural reasoning through two uncoupled phases. First, when new unstructured data enter the system, the \textit{PKG Constructor} performs a neural-to-symbolic transduction, transforming multimodal data into semantic triples to populate the user's PKG. Second, when the user interacts with the system, the reasoning is conducted directly on the PKG. This phase actively combines symbolic graph operations with neural RAG grounded in the graph's topology. Differently from standard Type~3 Neuro-Symbolic,  EpisTwin leverages both symbolic methods and neural models at inference time to extract new relevant and contextual information (see~\Cref{sec:agent_workflow}). Indeed, in the initial neural-to-symbolic transduction, some crucial information may be lost due to the context-unaware nature of the operation. 
Once the system generates an answer for the user query, EpisTwin \textit{Core Agent} judges the quality of the response. If the answer is unsatisfactory and the reasoning involves knowledge deriving from non-textual content (e.g., images, audio), the Core Agent triggers the \textit{Fallback agent}.
This approach may cover a wide range of different modalities. Without loss of generality, EpisTwin considers only images in this category. 
The Fallback Agent exploits its \textit{Online Deep Visual Refinement} tool, leveraging neural retrieval models to select the most relevant images, referenced in the PKG, to extract precise, context-aware attributes with neural vision models.
Additionally, the reliance on symbolic data ensures that deleting a node in the graph permanently excises the information, a guarantee almost impossible with purely neural or dense-vector approaches (see the recent research about \textit{unlearning}~\cite{DBLP:journals/corr/abs-2103-14991,DBLP:conf/aaai/YangH0025}).
To evaluate our framework, in the scarcity of open benchmarks for personal AI, we introduce \textit{PersonalQA-71-100}, a curated, privacy-preserving dataset, comprising $71$ heterogeneous personal data streams and $100$ queries with ground-truth answers. 
Evaluation is conducted using an LLM-as-a-Judge protocol~\cite{liuetal2023g,lee-etal-2025-checkeval}, enabling automated assessment of semantic correctness across multi-step and multi-modal reasoning tasks. Results show that EpisTwin consistently achieves satisfactory performance, even in scenarios requiring multimodal integration.
In summary, this paper makes the following contributions:
\begin{itemize}

    \item \textbf{Type~3 Neuro-Symbolic Agentic Framework:} EpisTwin unifies PKG construction and agentic reasoning to enable multi-hop question answering over heterogeneous personal data.

    \item \textbf{Active Visual-Symbolic Inference:} Our \textit{Online Deep Visual Refinement} tool recovers details lost during the transduction phase by leveraging symbolic knowledge for query-conditioned, on-the-fly visual analysis.


    \item \textbf{Novel Multimodal Benchmark:} We present \textit{PersonalQA-71-100}, a multimodal dataset of fragmented, realistic personal data (e.g., notes, calendar, images) paired with questions and gold-standard answers.
    

    \item \textbf{Verifiable Performance Analysis:} We adopt a ``LLM-as-a-Judge'' evaluation protocol to quantify reasoning capabilities. We demonstrate that EpisTwin handles the complexity of cross-domain, multi-step user queries.
\end{itemize}

\section{Related Work}
\label{sec:related_work}
Generative knowledge extraction and graph-based retrieval have seen significant progress, yet current paradigms struggle to connect unstructured perception and symbolic reasoning on personal data. In this section, we analyze prior work on knowledge graph construction, reasoning, personal AI, and data sovereignty to highlight the limitations that motivate an agentic, graph-grounded approach.

\paragraph{Knowledge Graph Construction.} 
Large Language Models (LLMs) have enabled triple extraction from unstructured text~\cite{DBLP:conf/emnlp/WeiH0K23}, but recent frameworks (e.g., RAKG~\cite{DBLP:journals/corr/abs-2504-09823}) remain predominantly unimodal. They fail to integrate structured logs and visual data essential for a digital twin. Moreover, Human-in-the-Loop strategies~\cite{DBLP:conf/esws/SchroderJ022} limit the scalability required for real-time personal intelligence.

\paragraph{Personal Knowledge Graphs.} Personal Knowledge Graphs differ from general KGs, necessitating strict data sovereignty, subjective context, and continuous evolution~\cite{BalogK19,DBLP:journals/aiopen/SkjaevelandBBLL24}. The current landscape is characterized by domain-specific fragmentation. Conversational agents~\cite{DBLP:conf/slt/LiTHL14} and temporal models~\cite{DBLP:conf/hsi/SansenCJTBH24} lack a holistic view of the user's digital ecosystem. Specialized implementations in healthcare~\cite{DBLP:conf/iscc/LiPH23}
and education~\cite{DBLP:conf/www/Ilkou22} are constrained by static ontologies, and restricted to specific domains.

\paragraph{Graph-based Reasoning.} To ensure logical validity, frameworks like Think-on-Graph (ToG)~\cite{DBLP:conf/iclr/SunXTW0GNSG24} and FiDeLiS~\cite{DBLP:conf/acl/SuiHLHWH25} constrain LLM generation to valid graph traversals via beam search or logical entailment. However, these methods rely on a static, complete knowledge base. Graph Retrieval-Augmented Generation (GraphRAG) architectures aim to overcome the ``local myopia'' of vector-based RAG~\cite{DBLP:conf/nips/LewisPPPKGKLYR020} by exploiting structural dependencies. Systems such as QA-GNN~\cite{DBLP:conf/naacl/YasunagaRBLL21} and recent PKG-based assistants~\cite{DBLP:conf/sigir/LiuDZGWW24,11181301} demonstrate that grounding generation in a graph enhances multi-hop reasoning.


\paragraph{Personal AI and Data Sovereignty.} Existing systems are functionally narrow, with PKG adoption limited to specific domains~\cite{DBLP:journals/fdgth/CarbonaroMNM24}. Moreover, compliance with tightening regulations~\cite{AIAct,piplchina24,cpra24} requires deterministic data removal, a non-trivial challenge for neural unlearning~\cite{DBLP:journals/tist/NguyenHRNLYN25}. EpisTwin addresses this by strictly decoupling reasoning from storage, ensuring data sovereignty is an architectural guarantee rather than a probabilistic behavior.

\section{Problem Definition and Preliminaries}
User information is inherently fragmented, scattered across isolated sources that range from structured calendar logs to unstructured document repositories. This compartmentalization obscures the latent semantic connections required for intelligent assistance. 
Formally, let $\mathcal{U}$ denote the set of users and $\mathcal{S} = \{S_1, S_2, \dots, S_n\}$ be the set of available \textbf{Data Sources}. We define the atomic unit of information generated by these sources as an \textit{Information Object}.
\
\begin{definition}[Information Object]
\label{def:information_object}
An \textbf{Information Object} is the atomic unit of digital information, formally defined as a tuple $\iota = (\sigma, \mu, c)$, where:
\begin{itemize}
    \item $\sigma \in \mathcal{S}$ denotes the \textbf{Source Provenance}, where $\mathcal{S}$ represents the finite set of distinct data silos (e.g., \textit{Calendar}, \textit{Gallery}) from which the data originates;
    \item $\mu$ represents the \textbf{Structured Metadata} (e.g., timestamps, file paths), defined as a set of explicit attributes, key-value pairs, such that $\mu \models \mathcal{M}_\sigma$, where $\mathcal{M}_\sigma$ denotes the schema governing valid metadata for the provenance $\sigma$;
    \item $c \in \mathcal{C} \cup \{\emptyset\}$ denotes the optional \textbf{Unstructured Payload}, where $\mathcal{C}_\sigma$ represents the domain of raw data made available by $\sigma$ (e.g., image tensors $\mathbb{R}^{H \times W \times 3}$).
\end{itemize}

\end{definition}

\begin{example}
Let us imagine a photo of Sagrada Familia in the photo gallery as an Information Object $\iota_\text{ph} = (\sigma_\text{ph}, \mu_\text{ph}, c_\text{ph})$, where $\sigma_\text{ph}$ is the \textit{Photos App} on the user's device, $\mu_\text{ph}$ contains the metadata key-value pairs (e.g. time: 10:05, date: 12-Jun-2025) and $c_\text{ph}$ represents the raw RGB pixel data.
\end{example}

The \textit{Information Objects} that are \textbf{owned by or pertaining to} a user $u$ are considered their \textit{Personal Knowledge}.

\begin{definition}[Personal Knowledge]
\label{def:personal_knowledge}
The \textbf{Personal Knowledge} $\mathcal{K}_u$ for a specific user $u \in \mathcal{U}$ is the heterogeneous set of all Information Objects associated with $u$, $\mathcal{K}_u = \{ \iota_1, \iota_2, \dots, \iota_m \}$,
where each $\iota_i$ is an Information Object generated by a source $\sigma \in \mathcal{S}$.
\end{definition}

\begin{remark}
\label{rem:pkg_constraints}
Associating an object $\iota$ with user $u$ implies that $\iota$ is either explicitly generated by $u$ (ownership) or semantically refers to $u$ (participation), making $\mathcal{K}_u$ \emph{user-centric}.
\end{remark}

The core challenge lies in mapping this unstructured and disconnected set $\mathcal{K}_u$ into a coherent epistemic structure that enables a reasoning agent; this implies two distinct but interdependent sub-challenges:

\begin{enumerate}  
    \item \textbf{Personal Knowledge Graph (PKG) Construction}: The process to map the \textit{Information Objects} into a PKG, converting implicit relations into explicit triples (\textit{PKG
    Population} in~\citet{DBLP:journals/aiopen/SkjaevelandBBLL24});
    \item \textbf{Personalized Answering}: The inference process to answer a natural language query exploiting the knowledge in the PKG (\textit{PKG Utilization} in~\citet{DBLP:journals/aiopen/SkjaevelandBBLL24}). Here, this step adopts a neuro-symbolic approach to cross-reference the heterogeneous user data, operation impossible with the sole symbolic reasoning.
\end{enumerate}

\subsection{Personal Knowledge Graph (PKG)}
\label{sec:pkg_preliminaries}
General Knowledge Graphs (KGs) serve as repositories for universal truths and globally significant entities. \textit{Personal Knowledge Graphs} (PKGs) represent the subjective context of a single individual, including only entities and relationships relevant to that specific user. \citet{BalogK19} originally characterized the PKG through a structural lens, defining it as a ``spiderweb" layout where the user acts as the central node connected to every other entity. \citet{DBLP:journals/aiopen/SkjaevelandBBLL24} subsequently argued that the defining feature is not topology but data sovereignty, viewing the PKG as a resource where the user is the sole administrator with exclusive access rights. We synthesize these complementary perspectives in our framework. The proposed model enforces administrative ownership of~\citet{DBLP:journals/aiopen/SkjaevelandBBLL24} while simultaneously retaining the user as the central anchor for all stored information as required by~\citet{BalogK19}.



\begin{definition}[Personal Knowledge Graph]
A Personal Knowledge Graph $\mathcal{G}$ is defined by a 
tuple $(
\mathcal{N}, \mathcal{R}, \mathcal{T}
)$, where: $\mathcal{N} = \mathcal{E} \sqcup \mathcal{L} \sqcup \mathcal{B}$ is the set of nodes, composed of the disjoint union of (i) Entities $\mathcal{E}$ (containing the special entity $u$ representing the owner of the Knowledge Graph), (ii) Literals $\mathcal{L}$, and (iii) Blank Nodes $\mathcal{B}$; $\mathcal{R}$ is a set of relations, $\mathcal{T} \subseteq (
\mathcal{E} \cup \mathcal{B}) \times \mathcal{R} \times (\mathcal{N}/\{u\})$ is a set of triples, where each $(h, r, t) \in \mathcal{T}$ represents a fact where $h$ is the head, $r$ is the relation, and $t$ is the tail. Structurally, $\mathcal{G}$ is 
rooted at $u$, and satisfies the reachability constraint: $\forall v \in \{t \mid \exists h, r : (h, r, t) \in \mathcal{T}\}$, there is a path from $u$ to $v$, denoted by $u \rightsquigarrow v$. 
\end{definition}

\subsection{KG Construction from Text}
\label{sec:kgc_preliminaries}

In this study, we leverage the generative capabilities of LLMs for Knowledge Graph Construction (KGC)~\cite{10.1145/3618295}, thus enabling the automated transformation of natural language into structured symbolic representations. 
For the unstructured non-textual content $c$, we assume there exists an operator that transforms an unstructured content $c$ into its textual representation $\hat{c}$ (later defined in~\Cref{sec:pkg_population}).


\begin{definition}[KG Construction Function]
\label{def:kgc}
Let $\hat{c}$ be the text derived from the unstructured content of an Information Object $\iota$. We define the KG Construction Function $f_{\text{KGC}}$ as a mapping that extracts a set of semantic triples $\mathcal{T}_{c}$ from a sequence generated by an LLM parameterized by $\theta$. Formally, $\mathcal{T}_{c} = \xi(y)$ where $y$ is the token sequence sampled from the conditional distribution of the LLM $y \sim P_{\theta}(\cdot \mid \hat{c})$, and $\xi$ is a parsing function that maps $y$ to a set of triples $\{(h, r, t)\} \subseteq \mathcal{E} \times \mathcal{R} \times (\mathcal{E}\cup\mathcal{L})$.
%
%
\end{definition}

The extraction is not a deterministic parsing task, but a dynamic inference process where $f_{\text{KGC}}$ takes into account the specific semantic nuances of the source $\delta$ to capture the heterogeneity typical of personal unstructured data.

\subsection{Thematic Context Extraction via Community Detection}
\label{sec:community_detection}
PKGs may lack explicit triples for semantically cohesive concepts (e.g., a link between an \textit{Event} and its reminder \textit{Alarm}).
To uncover these latent associations, we employ a \textit{Community Detection Algorithm}~\cite{Fortunato_2010}.

\begin{definition}[Community Structure]
Given the Personal Knowledge Graph $\mathcal{G} = (\mathcal{N}, \mathcal{R}, \mathcal{T})$, a \textit{Community Structure} is defined as a collection of clusters $\mathcal{P} = \{P_1, P_2, \dots, P_k\}$, with each cluster $P \subseteq \mathcal{N}$.
The quality of this thematic collection is evaluated via \textit{Modularity} objective $Q$~\cite{Fortunato_2010}.
Maximizing $Q$ yields communities where the intra-cluster topological density significantly exceeds that of a random null model, implying strong thematic coherence.
\end{definition}

\paragraph{Leiden Algorithm.}

The algorithm optimizes $Q$ through an iterative process composed by \textit{local moving} (greedy node reassignment), \textit{refinement} (splitting sub-optimal clusters to ensure connectivity), and \textit{aggregation} (coarsening the network).
To compute $P$, we select Leiden~\cite{Traag_2019} over the classic Louvain~\cite{Blondel_2008} method because it guarantees the \textit{connectivity} of the resulting communities. In a personal knowledge context, a disconnected community (e.g., a ``Work" cluster containing two unrelated sub-graphs) would represent a failure in semantic disambiguation. Each identified community is reified in the graph as a new entity connected to all the members of the community.

\subsection{Graph Retrieval-Augmented Generation}
\label{sec:graphrag_preliminaries}

We adopt \textit{Graph Retrieval-Augmented Generation} (GraphRAG)~\cite{DBLP:conf/nips/LewisPPPKGKLYR020,Edge} as the retrieval mechanism to enable reasoning over the PKG. Unlike standard RAG, which treats the LLM as a stochastic processor of disjoint text chunks, GraphRAG leverages the structural dependencies of the knowledge graph to capture relational context.
Following \citet{Peng2025}, the generation process decomposes into two stages: a retriever selecting relevant subgraphs, and a generator that synthesizes responses conditioned on this evidence. Given a user query $q$ and the PKG $\mathcal{G}$, the answer $a$ is generated through the process $f_{G\text{-}RAG}(q,\mathcal{G};\theta,\phi)$ defined as the sampling $a \sim P_\phi(a|q, \mathcal{G^*}) \cdot P_\theta(\mathcal{G^*}|q, \mathcal{G})$, where:
\begin{itemize}
    \item $P_\theta(\mathcal{G^*}|q, \mathcal{G})$ is the retriever, selecting the optimal subgraph $\mathcal{G^*} \subseteq \mathcal{G}$ via topological operators (e.g., community clusters, ego-networks);
    \item $P_\phi(a|q, \mathcal{G^*})$ is an LLM with parameters $\phi$ generating the response $a$ conditioned on the retrieved subgraph.
\end{itemize}

\subsection{LLM-based Agents}
\label{sec:agents_preliminaries}
Static execution flows often lack the flexibility to disentangle ambiguous user queries or synthesize insights from complex sets of triples in the PKG. To bridge this gap, we employ \textit{Agents} powered by Large Language Models (LLMs)~\cite{YaoZYDSN023,SchickDDRLHZCS23} to orchestrate adaptive reasoning over information-dense contexts. 
We define the agent $\mathcal{A}$ as a tuple $(\pi_\theta, \Sigma, \Lambda)$, where $\pi_\theta$ is a decision policy parameterized by the LLM, $\Sigma$ is the dynamic state space, and $\Lambda$ is a set of executable functional operators.
The reasoning proceeds as a discrete control loop. At step $t$, the agent observes the current state $s_t \in \Sigma$, defined as $s_t = (q, \mathcal{H}_t)$,
where $q$ is the initial query and $\mathcal{H}_t$ represents the history of thoughts and observations (the \textit{cognitive trajectory}).
Based on $s_t$, the policy samples an action $a_t \sim \pi_\theta(a | s_t)$, where $a_t \in \Lambda$.

\section{Epistemic Twin Constructor}
\label{sec:epistwin_constructor}
The Epistemic Twin Constructor implements the \textit{
PKG population} task described by \citet{DBLP:journals/aiopen/SkjaevelandBBLL24}. It projects each Information Object $\iota = (\sigma, \mu, c) \in \mathcal{K}_u$ into the user's PKG. The graph $\mathcal{G}_u$ is initialized with $\mathcal{E} = \{u\}$ and iteratively expanded when each $\iota$ is processed.
 
\subsection{PKG Population}
\label{sec:pkg_population}
\begin{figure}[t]
    \centering
    \includegraphics[width=0.70\linewidth]{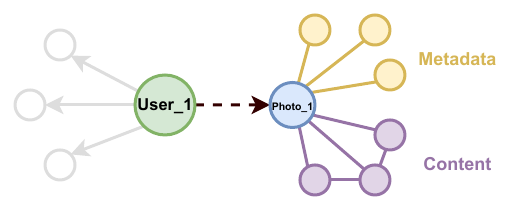} 
    \caption{
    PKG population when the Information Object is a photo: triples are extracted from both metadata and visual content.}
    \label{fig:aggregation}
\end{figure}

We model the \textbf{PKG Population Task} as an update function $\Psi$ that transitions the graph from state $\mathcal{G}^k_u$ to $\mathcal{G}^{k+1}_u$ by merging the current graph with the new subgraph, containing the triples extracted from the information object $\iota^{k+1}$:
\begin{equation}
    \mathcal{G}^{k+1}_u = \Psi(\mathcal{G}^k_u, \iota^{k+1}) = \mathcal{G}^k_u \oplus  \left( \Phi_{\mathcal{M}}(\mu) \cup \Phi_{\mathcal{C}}(c)\right)
\end{equation}
where $\Phi_{\mathcal{M}}$ represents a deterministic translation of structured metadata $\mu$ into a knowledge graph, $\Phi_{\mathcal{C}}$ denotes the creation of a knowledge graph from the unstructured content $c$, and $\oplus$ is a merge operator which joins the newly computed subgraph to $\mathcal{G}_u^k$. See a visualization of this process in Figure~\ref{fig:aggregation}.

\subsubsection{Metadata triples extraction - $\Phi_{\mathcal{M}}$ function}
Let $\mathbb{M} = \mathcal{K} \times \mathcal{V}$ be the space of all possible metadata entries, and let $\mathbb{T} = \mathcal{N} \times \mathcal{R} \times (\mathcal{E}\cup\mathcal{L})$ be the universe of valid symbolic triples (consistent with the domain of $\mathcal{T}$ in~\Cref{sec:pkg_preliminaries}).
We formulate $\Phi_{\mathcal{M}}$ as the functional composition $\Phi_{\mathcal{M}} = \Phi_{\mathcal{M}_2} \circ \Phi_{\mathcal{M}_1}$. The first component $\Phi_{\mathcal{M}_1}: 2^{\mathbb{M}} \rightarrow 2^{\mathbb{T}}$, is a mapping from the power set of metadata to the power set of triples:
\begin{equation}
    \Phi_{\mathcal{M}_1}(\mu) = \Big\{ \big(n_\iota, \rho(k), \lambda(v)\big) \mid (k, v) \in \mu \Big\}
\end{equation}
where $n_\iota \in \mathcal{N}$ is the node for $\iota$, $\rho$ maps keys to predicates, and $\lambda$ casts values into entities or literals. Finally, $\Phi_{\mathcal{M}_2}$ transforms the triples extracted from $\iota$ into a knowledge graph.
\subsubsection{Unstructured Triples Extraction: the $\Phi_{\mathcal{C}}$ function}

Let $\mathcal{C}$ denote the space of unstructured content, including textual and visual inputs.
The operator $\Phi_{\mathcal{C}} : \mathcal{C} \rightarrow \mathcal{G}$ maps a content item $c \in \mathcal{C}$ to a knowledge graph by extracting symbolic facts and linking them to their source node $n_\iota$.

\paragraph{Textual Normalization.}
To enable uniform symbolic processing across modalities, we first define a normalization operator
$\eta : \mathcal{C} \rightarrow \mathcal{C}_{\text{text}}$,
where $\mathcal{C}_{\text{text}}$ is the space of textual representations. Without loss of generality, in this paper, we limit the scope of non-textual unstructured content to images. The operator $\eta$ is defined as
\[
\eta(c) =
\begin{cases}
c & \text{if } c \in \mathcal{C}_{\text{text}}, \\
\tau(c) & \text{if } c \in \mathcal{C}_{\text{vis}},
\end{cases}
\]
where $\mathcal{C}_{\text{vis}}$ denotes the space of visual content and $\tau$ is a captioning operator.
For visual inputs, $\tau$ is implemented using a multimodal language model parameterized by $\omega$, which induces a conditional distribution over textual descriptions
$\tau(c) \sim P_{\phi}(\cdot \mid c, \textit{prompt}_{\text{vis}})$,
where $\textit{prompt}_{\text{vis}}$ is a dense captioning prompt. In practice, a single realization is used as the normalized textual representation.

\paragraph{Triple Extraction and Graph Integration.}
Given a textual representation $\hat{c} = \eta(c)$, a knowledge graph construction operator $f_{\text{KGC}}$, defined in~\Cref{sec:kgc_preliminaries}, maps text to a set of symbolic triples $\mathcal{C}_{\text{text}} \rightarrow 2^{\mathcal{E} \times \mathcal{R} \times (\mathcal{E}\cup\mathcal{L})}$.

The triples extracted with $f_{\text{KGC}}$ are transformed into a knowledge graph $\mathcal{G}_{c} = \Phi_{\mathcal{C}}(c)$ and integrated into the PKG via $\Psi\!\left(\mathcal{G}, \mathcal{G}_{c}\right)$. The sequential application of captioning and symbolic extraction,
$c \xrightarrow{\;\eta\;} \hat{c} \xrightarrow{\;f_{\text{KGC}}\;} \mathcal{G}_{c}$,
constitutes a \emph{Visual–Symbolic Transduction} process, whereby raw perceptual inputs are projected into the PKG.

\subsection{Community Detection as PKG Post-processing}
\label{subsec:post_processing}

\begin{figure*}[t]
    \centering
    \includegraphics[height=0.137\linewidth]{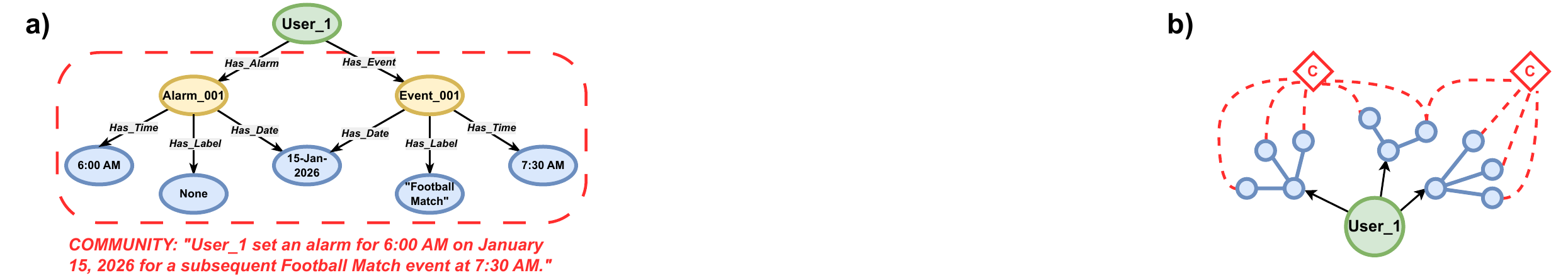} 
    \caption{
    Communities over the PKG: (a) The topologically disjoint entities ``Alarm" and ``Football Match Event" could be grouped into a shared community that reveals an implicit consequentiality, improving reasoning. (b) A macroscopic visualization of a PKG populated by entities, relationships, and thematic communities.}
    \label{fig:communities}
\end{figure*}

As the PKG is populated with high-density, granular triples, the resulting topology often suffers from structural dispersion, hindering effective global sensemaking, particularly as the graph scales. To mitigate this, we implement a \textit{post-processing routine} for detecting thematic communities, executed after each update of $\mathcal{G}_u$. Figure~\ref{fig:communities} shows how thematic communities could be integrated over the PKG topology.
We utilize the Leiden algorithm to identify a \textit{Community Structure} $\mathcal{P} = \{P_1, \dots, P_k\}$ (as defined in \Cref{sec:community_detection}), optimizing for the modularity objective. Then we \textit{reify} them into the graph as distinct nodes that serve as high-level access points for reasoning. 
We extend the original PKG topology $\mathcal{G}_u = (\mathcal{N}, \mathcal{R}, \mathcal{T})$ to an augmented graph $\mathcal{G}'_u = (\mathcal{N}', \mathcal{R}', \mathcal{T}')$ via the following transformation:
\begin{align}
    \mathcal{N}' &= \mathcal{N} \cup \{ n_{P} \mid P \in \mathcal{P} \} \\
    \mathcal{R}' &= \mathcal{R} \cup \{\texttt{in\_community}\} \\
    \mathcal{T}' &= \mathcal{T} \cup \bigcup_{P \in \mathcal{P}} \{ (n, \texttt{in\_community}, n_{P}) \mid n \in P \}
\end{align}
\noindent where $n_{P}$ is a \textit{Community Node} denoting the cluster $P \subseteq \mathcal{N}$. 

\paragraph{Generative Summarization.} 
To enable semantic reasoning over these structures, each community node $n_{P}$ is enriched with a generated summary $S_{P}$, which synthesizes the information contained within the subgraph induced by the community.
We define the induced subgraph topology $\mathcal{T}_{P}$ as the subset of triples strictly contained within the cluster:
\begin{equation}
    \mathcal{T}_{P} = \{ (h, r, t) \in \mathcal{T} \mid h \in P \land t \in P \}
\end{equation}

To generate the summary, we introduce the \textbf{Generative Summarization Operator} $f_\text{SUM}$. Since the underlying mechanism is a Large Language Model, the output is obtained by sampling from a conditional distribution. Formally:
\begin{equation}
    S_{P} = f_\text{SUM}(P, \mathcal{T}_{P}) \quad \text{where} \quad S_{P} \sim P_\gamma( \cdot \mid \text{lin}(P, \mathcal{T}_{P}) )
\end{equation}
Here, $P_\gamma$ represents the LLM parameterized by $\gamma$, and $\text{lin}(\cdot)$ is a linearization function that serializes the graph elements (nodes and triples) into a textual prompt context.

\section{EpisTwin Reasoning Engine}
\label{sec:reasoning_architecture}

The previous sections detailed the construction of the PKG. However, the static graph alone is insufficient for holistic sensemaking. The \textit{EpisTwin Reasoning Engine} is designed not as a static retrieval pipeline, but as a \textit{Cooperative Neuro-Symbolic Orchestrator}. 
Rather than a rigid hierarchy, the system operates as a dynamic state machine that transitions between \textit{symbolic reasoning} (traversing $\mathcal{G}_u$) and \textit{neural perception} (analyzing raw modalities) based on the epistemic confidence of the current state.
This architecture ensures that the system maximizes the utility of the verifiable graph structure while adaptively grounding reasoning in raw data when symbolic representations are sparse or ambiguous. The decision to exploit agentic AI is motivated by the potential to enable additional specialized functionalities, such as Health Assistance or Point-of-Interest (PoI) recommendation (e.g., \textit{``I am in Paris, recommend a restaurant based on my past dining preferences''}).

\subsection{Agentic Orchestration}
\label{sec:agent_workflow}

We model the reasoning process as a sequential decision-making problem. The \textbf{Core Agent} $\Delta_{\text{Core}}$ acts as the primary controller, governed by a policy $\pi_\theta$ parameterized by a Large Language Model.
At time step $t$, the system state is defined as $s_t = (q, \mathcal{H}_{t})$, where $q$ is the user query and $\mathcal{H}_{t}$ represents the reasoning trajectory (Chain-of-Thought).
The agent selects an action $a_t \sim \pi_\theta(a | s_t)$ from a composite action space $\Lambda$. This space unifies symbolic tools (e.g., community detection, ego-network expansion) and delegation triggers.
To ensure response fidelity, the workflow integrates an \textit{Epistemic Verification Module}, $f_{\text{VAL}}$, which functions as a self-reflection operator. After deriving a candidate reasoning step, the system evaluates semantic sufficiency $v_t \sim P_\phi(v | q, \mathcal{H}_t \in \{\texttt{Sufficient}, \texttt{Insufficient}\})$.
If $v_t = \texttt{Insufficient}$, implying that the symbolic graph lacks the granularity to answer $q$, the policy $\pi_\theta$ shifts the reasoning mode. It triggers the \textbf{Fallback Agent} $\Delta_{\text{FB}}$ not as a subordinate, but as a specialized \textit{neural co-routine} designed to handle unstructured modalities, effectively expanding the state space beyond the graph's limits.

\subsection{Bridging the Gap: Online Visual Refinement}
\label{sec:visual_refinement}

A fundamental challenge in Personal AI is related to the intrinsic contextual nature of user queries. Therefore, the initial transduction $\Phi_C$ (\Cref{sec:pkg_population}), compressing high-dimensional sensory data into discrete triples without any context, may inevitably discard content useful in specific contexts.
To resolve this, we introduce \textbf{Online Deep Visual Refinement} ($t_{\text{VIS}}$). This tool is invoked when the self-reflection step $v_t$ indicates a deficit in symbolic information regarding visual entities.
The refinement process operates via \textit{query-driven re-grounding}. Let $\mathcal{E}_q \subset \mathcal{G}_u$ be the set of Information Objects entities topologically relevant to the query (e.g., photos linked to the query topic). The $t_{\text{VIS}}$ tool executes a dual-phase neural lookup:
\begin{enumerate}
    \item \textbf{Contextual Fetching:} It retrieves the raw unstructured payload $c$ (e.g., original image tensors) associated with nodes in $\mathcal{E}_q$.
    \item \textbf{Neural VQA Injection:} It employs a Multimodal LLM $M_{\text{vis}}$ to re-analyze the raw content of the information objects relevant to $q$; $a_{\text{vis}} = \text{Agg}\left( \{ M_{\text{vis}}(q, c) \mid c \in \text{payload}(\mathcal{E}_q) \} \right)$
\end{enumerate}
The output $a_{\text{vis}}$ is a natural language synthesis of visual evidence. Crucially, the extracted insights are treated as \textit{ephemeral context}, injected into the reasoning history $\mathcal{H}_t$ for the current session only. This prevents the permanent pollution of the curated PKG with transient, query-specific noise, maintaining the graph as a deterministic source of truth while allowing for flexible, on-demand neural perception.

\section{Experimental Evaluation}
\label{sec:exp_discussion}

The evaluation is designed to assess the system's capabilities in retrieving, connecting, and reasoning on multimodal personal data within a controlled setting. To the best of our knowledge, this is one of the first studies in this direction. 

\subsection{The \textit{PersonalQA-71-100} Benchmark}
\label{subsec:setup}
 


To validate the proposed architecture, we designed \textit{PersonalQA-71-100}, a synthetic benchmark composed of two synchronized collections (i.e., data and Question-Answer pairs). 
The first comprises \textbf{71 synthetic Information Objects} drawn from seven distinct sources (Calendar, Alarm, Photos, Note, Documents, Phone, and Contacts; see \Cref{tab:combined_stats}). The other consists of \textbf{100 query-answer samples} specifically authored to probe the system's reasoning limits, structured as triplets $(q, a_{target}, a_{ET})$ where the ground truth $a_{target}$ only contains strictly needed information to answer the question. The EpisTwin-generated answer $a_{ET}$ is released as well.
The benchmark is designed to test the architecture across three core cognitive dimensions: \textbf{Temporal Reasoning}, evaluating the resolution of indexical expressions and scheduling conflicts; \textbf{Cross-Source Reasoning}, challenging the synthesis of fragmented information across heterogeneous applications (Table~\ref{tab:combined_stats}); and \textbf{Fact Retrieval}, assessing precision in extracting granular details or summarizing recurring patterns found in the data.
To guarantee a robust assessment of temporal logic, 
the querying time is fixed to $T_{global} = \text{2025-Sep-01 at 13:00}$,
and the information objects were created with a creation date previous to $T_{global}$.

\begin{table}[t!]
    \centering
    \small 
    \setlength{\tabcolsep}{1.5pt} 
    
    \newcolumntype{R}{>{\raggedleft\arraybackslash}X}

    \begin{tabularx}{\columnwidth}{@{}lRRRRRRR@{}} 
        \toprule
        \textbf{App} & \textbf{Events} & \textbf{Images} & \textbf{Notes} & \textbf{Docs} & \textbf{Calls} & \textbf{Alarms} & \textbf{Contacts} \\
        \midrule
        N. & 20 & 15 & 15 & 9 & 6 & 4 & 2 \\
        \bottomrule
    \end{tabularx}
    
    \vspace{5pt} 

    \begin{tabularx}{\columnwidth}{@{}lRRRR@{}} 
        \toprule
         & \textbf{1 App} & \textbf{2 Apps} & \textbf{3 Apps} & \textbf{4 Apps} \\
        \midrule
        Distribution & 63 & 32 & 4 & 1 \\
        \bottomrule
    \end{tabularx}

    \caption{Data distribution in \textit{PersonalQA}. \textbf{Top:} Distribution of questions across data sources. \textbf{Bottom:} Distribution of questions based on the number of data sources involved in the reasoning process.}
    \label{tab:combined_stats}
\end{table}


    


\subsection{Implementation Details}
The EpisTwin architecture is instantiated as a modular system. We employ \textbf{Neo4j}~\footnote{https://neo4j.com/}
as the underlying graph database, leveraging its native hybrid indexing to support both Graph Retrieval ($f_{\text{G-RAG}}$, \Cref{sec:graphrag_preliminaries}) via \textbf{GraphRag}~\cite{Edge} and Knowledge Graph Construction (\Cref{sec:kgc_preliminaries}) via \textbf{LLM Graph Builder}~\footnote{https://neo4j.com/labs/genai-ecosystem/llm-graph-builder/}. The inference logic is distributed across specialized models: \textbf{Qwen3-32B}~\cite{yang2025qwen3technicalreport} drives the Agents, \textbf{GPT-OSS}~\cite{openai2025gptoss} handles the symbolic triple extraction ($f_{\text{KGC}}$), \textbf{Gemini 2.5 Pro}~\cite{comanici2025gemini25} powers GraphRAG reasoning, while \textbf{LLaMA-4-maverick-17B-128e}~\cite{abdullah2025evolutionmetasllamamodels} is the MLLM which manages both visual captioning ($\tau$) and the Online Deep Visual Refinement tool ($t_\text{VIS}$), selected for its native capabilities in long-context visual reasoning.

\subsection{Evaluation Methodology: LLM-as-a-Judge}

To assess the performance of EpisTwin, we employ an automated, reproducible LLM-as-a-Judge framework~\cite{lee-etal-2025-checkeval,liuetal2023g}. 
Responses generated by our system ($a_{ET}$) were evaluated against the ground truth ($a_{target}$) to measure reasoning accuracy and retrieval precision.

\noindent \textbf{Judicial Panel Selection.}\label{subsubsec:panel_selection}
The evaluation relies on Open Source models in authoritative benchmarks, i.e., LLM Stats~\footnote{https://llm-stats.com},
to guarantee transparency, consistency, and reproducibility~\cite{liuetal2023g}. The panel comprises four state-of-the-art architectures: DeepSeek-V3.2~[\citeyear{deepseek2025v3}], Qwen3-32B~[\citeyear{yang2025qwen3technicalreport}], GPT-OSS-120B~[\citeyear{openai2025gptoss}], and Kimi K2 Instruct 0905~[\citeyear{kimiteam2025kimik2openagentic}]. This approach mitigates the single-judge evaluation biases and ensure a comprehensive assessment~\cite{Ahumanaicomparative}.

\noindent \textbf{Evaluation Strategy.} The evaluation process follows Prometheus~\cite{prometheusinducingfinegrainedevaluation} prompting strategy. This methodology enforces a Chain-of-Thought evaluation process and each judge is provided with (i) a prompt containing the user query ($q$), (ii) the system's generated response ($a_{ET}$), and (iii) the ground-truth answer ($a_{target}$). 
To minimize subjective bias, judges analyze the triplet $(q, a_{target}, a_{ET})$ and articulate step-by-step reasoning before assigning a score. The assessment relies on a 5-point \textit{Likert} scale ranging from \textit{1 (Irrelevant)} to \textit{5 (Ground-Truth Aligned)}.
However, considering that subtle score differences may occur from model to model~\cite{lee-etal-2025-checkeval}, we normalize scores to focus on semantic utility. Therefore, scores are assigned to three ordinal categories: \textit{Positive} (Scores 4-5) for accurate responses; \textit{Neutral} (Score 3) for relevant but incomplete outputs; and \textit{Negative} (Scores 1-2) for incorrect results. 


    

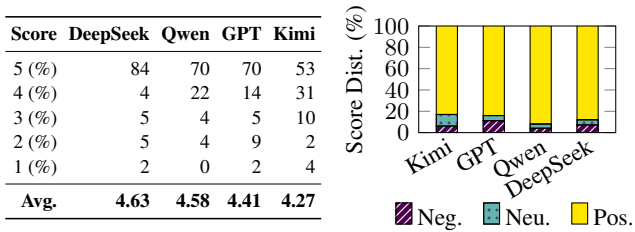
\begin{figure}[t!]
    \centering
    \begin{minipage}[c]{0.49\linewidth}
        \centering
        \renewcommand{\arraystretch}{1.15} 
        \setlength{\tabcolsep}{3pt}        
        
        \resizebox{\linewidth}{!}{
            \begin{tabular}{rrrrr}
            \toprule
            \textbf{Score} & \textbf{DeepSeek} & \textbf{Qwen} & \textbf{GPT} & \textbf{Kimi} \\
            \midrule
            5 (\%) & 84 & 70 & 70 & 53 \\
            4 (\%) & 4  & 22 & 14 & 31 \\
            3 (\%) & 5  & 4  & 5  & 10 \\
            2 (\%) & 5  & 4  & 9  & 2  \\
            1 (\%) & 2  & 0  & 2  & 4  \\
            \midrule
            \textbf{Avg.} & \textbf{4.63} & \textbf{4.58} & \textbf{4.41} & \textbf{4.27} \\
            \bottomrule
            \end{tabular}%
        }
    \end{minipage}%
    \hfill 
    \begin{minipage}[c]{0.49\linewidth}
        \centering
        \begin{tikzpicture}
        \begin{axis}[
            ybar stacked,
            bar width=8pt,
            width=\linewidth, 
            height=3.0cm,     
            axis line style={-},
            xtick style={draw=none},
            ymin=0, ymax=100,
            enlarge x limits=0.2,
            ylabel={Score Dist. (\%)},
            ylabel style={yshift=-0.4cm}, 
            label style={font=\fontsize{9pt}{10.8pt}\selectfont},
            tick label style={font=\fontsize{9pt}{10.8pt}\selectfont},
            symbolic x coords={Kimi, GPT, Qwen, DeepSeek},
            xtick=data,
            x tick label style={rotate=30, anchor=north east, align=right, inner sep=1pt},
            legend style={
                font=\fontsize{9pt}{10.8pt}\selectfont,
                at={(0.5,-0.6)}, 
                anchor=north,
                legend columns=-1,
                draw=none,
                fill=none,
                /tikz/every even column/.append style={column sep=0.2cm}
            },
        ]
        \addplot+[ybar, fill=violet!60!black, draw=black, postaction={pattern=north east lines, pattern color=white!30}] 
        table [x=model, y=0.0, col sep=semicolon] {images/data/score_distribution.csv};
        \addlegendentry{Neg.}
        
        \addplot+[ybar, fill=teal!60, draw=black, postaction={pattern=dots, pattern color=black!70}] 
        table[x=model, y=1.0, col sep=semicolon]{images/data/score_distribution.csv};
        \addlegendentry{Neu.}
        
        \addplot+[ybar, fill=yellow!90!orange, draw=black] 
        table[x=model, y=2.0, col sep=semicolon]{images/data/score_distribution.csv};
        \addlegendentry{Pos.}
        \end{axis}
        \end{tikzpicture}
    \end{minipage}
    
    \vspace{-0.1cm} 
    \caption{\textbf{Judicial Panel Evaluation.} \textbf{Left:} Distribution of scores assigned by LLM Judges on \textit{PersonalQA-71-100} to EpisTwin answers. \textbf{Right:} LLM judgments distribution after vote aggregation.}
    \label{fig:score_analysis}
\end{figure}

    
        

\noindent \textbf{Inter-rater Reliability.} To validate the robustness of our automated judicial panel, reporting the score aggregation provides an incomplete overview. Following the recent literature, we rely on a suite of metrics specifically proposed for these purposes. We calculate \textbf{Percentage Agreement} (\% Agr.)~\cite{Interraterreliability}  and \textbf{Quadratic Weighted Cohen’s Kappa} ($\kappa$)~\cite{Cohen1968Weighted} for ordinal reliability. However, \citet{Gwet2008Computing} noted that this metric is affected by a limitation: when there is high agreement and low variance, the best case, Kappa yields low scores. As a viable alternative metric, they propose \textbf{Gwet’s AC1} (AC1), that we included in the evaluation for completeness. 
Finally, we assess monotonic scoring trends using \textbf{Spearman’s Rank Correlation} ($\rho$)~\cite{Spearman1904Proof} and \textbf{Krippendorff’s Alpha} ($\alpha$)~\cite{Krippendorff2011ComputingKA} to evaluate global consistency. 

\section{Results Discussion}
\label{sec:discussion}
The empirical evaluation on \textit{PersonalQA-71-100} validates the core hypothesis of this work: a Neuro-Symbolic architecture, by decoupling reasoning from storage, can achieve high-fidelity ``global'' sensemaking over fragmented personal data. 

\subsection{Reasoning on Fragmented Heterogeneous Data}
As detailed in~\Cref{fig:score_analysis}, EpisTwin consistently achieves high performance across all four judge models, with mean scores ranging from \textbf{4.27 (Kimi)} to \textbf{4.63 (DeepSeek)}. Notably, the system received a \textbf{Positive rating (Score 4 or 5)} in \textbf{87\%} of the test cases (aggregated across judges, see~\Cref{fig:score_analysis}). 
The significance of this result lies not merely in the high accuracy, but in the \textbf{structural complexity} of the queries it resolves. As shown in~\Cref{tab:combined_stats}, the benchmark is designed to stress-test cross-silo reasoning, with a significant portion of queries requiring the agent to bridge up to four distinct applications (e.g., correlating \textit{Calendar} logs, \textit{Contact} details, and \textit{Photo} metadata) to derive a single answer. 
In standard vector-based RAG systems, performance typically degrades as the number of required ``hops'' increases due to context fragmentation~\cite{tao-etal-2025-saki}. In contrast, EpisTwin's robust performance confirms that the PKG successfully preserves the topological dependencies between these isolated sources. By grounding the agent in a unified semantic layer, the system effectively transforms a complex, multi-source retrieval problem into a traversable graph query, maintaining coherence where purely neural approaches often hallucinate.

\subsection{Robustness of the Evaluation Panel}
The validity of our results relies on the consistency of the ``LLM-as-a-Judge'' panel. We observe (Table~\ref{tab:pairwise_metrics}) strong inter-rater reliability, with \textbf{Gwet’s AC1 consistently exceeding 0.84} and \textbf{Percentage Agreement above 84\%}.
While variance-dependent metrics such as \textbf{Cohen’s $\kappa$} and \textbf{Krippendorff’s $\alpha$} yield lower values (avg. 0.65), this reflects the \textit{paradox of high agreement} common in skewed distributions~\cite{Gwet2008Computing}. Since the system performs correctly on the vast majority of queries, the scarcity of ``Negative'' labels artificially deflates these metrics. The significant divergence between AC1 (which is robust to trait prevalence) and $\kappa$ confirms that the judges function as a homogeneous entity. This statistical consensus demonstrates that the high scores reflect an objective alignment with the ground truth. 

\textit{Overall, the statistically robust consensus among state-of-the-art judge models proves that EpisTwin bridges the gap between fragmented personal data and agentic reasoning.}

\begin{table}[t!]
\small
\centering
\begin{tabular}{lrrrr}
\toprule
\textbf{Comparisons} & \textbf{ AC1} & \textbf{\% Agr.} & \boldmath$\kappa$ & \boldmath$\rho$ \\
\midrule
DeepSeek vs Qwen  &  0.92 & 93.0 & 0.73 & 0.70  \\
DeepSeek vs GPT   & 0.85 & 87.0 & 0.69 & 0.70  \\
DeepSeek vs Kimi  &  0.84 & 86.0 & 0.66 & 0.60  \\
Qwen vs GPT       &   0.84 & 86.0 & 0.54 & 0.60   \\
Qwen vs Kimi   &    0.86 & 88.0 & 0.70 & 0.69  \\
GPT vs Kimi    &     0.81 & 84.0 & 0.61 & 0.56  \\
\bottomrule
\end{tabular}%
\caption{
Statistical alignment using  Gwet's AC1 (AC1), Percentage Agreement (\%Agr.), Cohen's $\kappa$ and Spearman's $\rho$.}
\label{tab:pairwise_metrics}
\end{table}

\subsection{Limitations}
\label{subsec:limitations}
The prioritization of structural grounding and data sovereignty entails specific trade-offs.
 
 
\paragraph{Scalability and Model Dependency.} Symbolic transduction of long documents generates high-density subgraphs that can saturate the context window. This complexity necessitates high-capability LLMs for strict schema adherence, as smaller models ($<$10B) struggle with these constraints.
 
\paragraph{Latency and Engineering Trade-offs.} The architecture prioritizes quality and observability through a coordinated ecosystem of specialized components (Population vs. Agentic layers). This design introduces latency, as the Reasoning Engine demands multiple state-space hops to synthesize context, particularly when integrating multimodal data.

\section{Conclusion}
In this paper, we presented EpisTwin, a Type 3 neuro-symbolic architecture that addresses the critical fragmentation of personal digital footprints. By synthesizing a Personal Knowledge Graph (PKG) from heterogeneous sources and employing Online Deep Visual Refinement, we overcome the \textit{local myopia} of standard RAG systems. EpisTwin does not merely retrieve text; it actively investigates visual evidence and traverses temporal dependencies, mirroring human cognitive processes to answer complex queries. By grounding generation in a user-controlled PKG, we ensure that the ``right to be forgotten" is a technical reality rather than a probabilistic hope; deleting a node in EpisTwin is final and verifiable.
Beyond these structural benefits, EpisTwin represents a shift from passive retrieval to active sensemaking. By equipping the system with the agency to perform on-the-fly visual analysis--re-grounding symbolic entities in their raw pixel context only when necessary--we bridge the gap between the rigidity of knowledge graphs and the nuance of multimodal perception. The robust results achieved on our novel PersonalQA-71-100 benchmark validate that this dynamic interplay is essential for handling the complexity of real-world user data. 
We envision that the future of personal assistants lies in this cooperative neuro-symbolic paradigm, where AI is a powerful reasoning engine, but the user remains the undisputed owner of the map.

\section*{Ethical Statement}
This work prioritizes \textit{user agency} as an architectural imperative. \textbf{First,} EpisTwin operationalizes the \textit{Right to be Forgotten} (e.g., GDPR Article 17) through deterministic unlearning; unlike probabilistic neural networks or vector stores where data traces may linger, the deletion of a subgraph in our framework guarantees the immediate, verifiable excision of information. \textbf{Second,} we enforce a \textit{strict separation between reasoning and memory}. By utilizing frozen Large Language Models and treating user data strictly as ephemeral, inference-time context, we prevent the leakage of sensitive personal information into model weights. \textbf{Finally,} \textit{if employed local LMs, EpisTwin ensures physical sovereignty}, allowing users to confine their digital footprint to a completely isolated environment.
\bibliographystyle{named}
\bibliography{ijcai26}

\end{document}